\documentclass[letterpaper, 10 pt, journal, twoside]{IEEEtran}

\usepackage{graphicx}
\usepackage{units}
\usepackage{subcaption}
\usepackage[hidelinks]{hyperref}
\usepackage{tabulary}
\usepackage{tabularx}
\usepackage{multirow}
\usepackage{xcolor,colortbl}
\usepackage{booktabs}
\usepackage{mathtools}
\usepackage{amsmath}
\usepackage{algorithm}
\usepackage{algpseudocode}

\IEEEoverridecommandlockouts                              

\title{Autonomous Sweet Pepper Harvesting for Protected Cropping Systems}

\author{Christopher Lehnert$^{1}$, Andrew English$^{1}$, Christopher McCool$^{1}$, Adam W. Tow$^{1}$ and Tristan Perez $^{1}$

    \thanks{Manuscript received: September, 1, 2016; Revised December, 7, 2016; Accepted January, 1, 2017.}
    \thanks{This paper was recommended for publication by Editor Wan Kyun Chung upon evaluation of the Associate Editor and Reviewers' comments. *Here you can knowledge the organizations/grants which supported the work}
    \thanks{This research was supported by the Australian Research Council
    Centre of Excellence for Robotic Vision (project number CE140100016),
    Queensland University of Technology (QUT), Brisbane, Australia.}
    \thanks{$^{1}$Chris Lehnert, Andrew English, Christopher McCool, Adam W. Tow, Tristan Perez are with the School of Electrical Engineering and Computer Science, Queensland University of Technology, Brisbane, Australia
	{\tt\small c.lehnert@qut.edu.au}}%
	\thanks{Digital Object Identifier (DOI): see top of this page.}
}

\begin{document}

\maketitle

\begin{abstract}


In this paper we present a new robotic harvester (Harvey) that can autonomously harvest sweet pepper in protected cropping environments.
Our approach combines effective vision algorithms with a novel end-effector design to enable successful harvesting of sweet peppers.
Initial field trials in protected cropping environments, with two cultivar, demonstrate the efficacy of this approach achieving a 46\% success rate for unmodified crop, and 58\% for modified crop.
Furthermore, for the more favourable cultivar we were also able to detach 90\% of sweet peppers, indicating that improvements in the grasping success rate would result in greatly improved harvesting performance.


\end{abstract}

\section{INTRODUCTION}\label{sec:introduction}

\IEEEPARstart{T}{he} horticulture industry remains heavily reliant on manual labour, and as such is highly affected by labour costs. In Australia, harvesting labour costs in 2013-14 accounted for 20\% to 30\% of total production costs \cite{ABARE2014}. These costs along with other pressures such as scarcity of skilled labour and volatility in production due to uncertain weather events is putting profit margins for farm enterprises under tremendous pressure.

Robotic harvesting offers an attractive potential solution to reducing labour costs while enabling more regular and selective harvesting, optimising crop quality, scheduling and therefore profit. These potential benefits have spurred research in the use of agricultural robots for harvesting horticultural crops over the past three decades \cite{Kondoetall2011}. Autonomous harvesting is a particularly challenging task that requires integrating multiple subsystems such as crop detection, motion planning, and dexterous manipulation. Further perception challenges also present themselves, such as changing lighting conditions, variability in crop and occlusions. A recent survey of 50 projects in robotic harvesting of horticulture crops \cite{Bac2014} highlighted that over the past 30 years of research, the performance of automated harvesting has not improved substantially despite huge advances in sensors, computers, and machine intelligence. If robotic-crop harvesting is to become a reality, we believe there are three key challenges that must be addressed: 
\begin{enumerate}
    \item \textit{Detection:} determining the presence/location of each crop 
    \item \textit{Grasp selection:} determining the 3D pose and shape of each crop and selecting appropriate grasping and/or cutting points.
    \item \textit{Manipulation:} detaching the crop from the plant without harming the crop or plant.   
\end{enumerate}

 


 \begin{figure}
   	\centering
 	\includegraphics[width=0.7\linewidth]{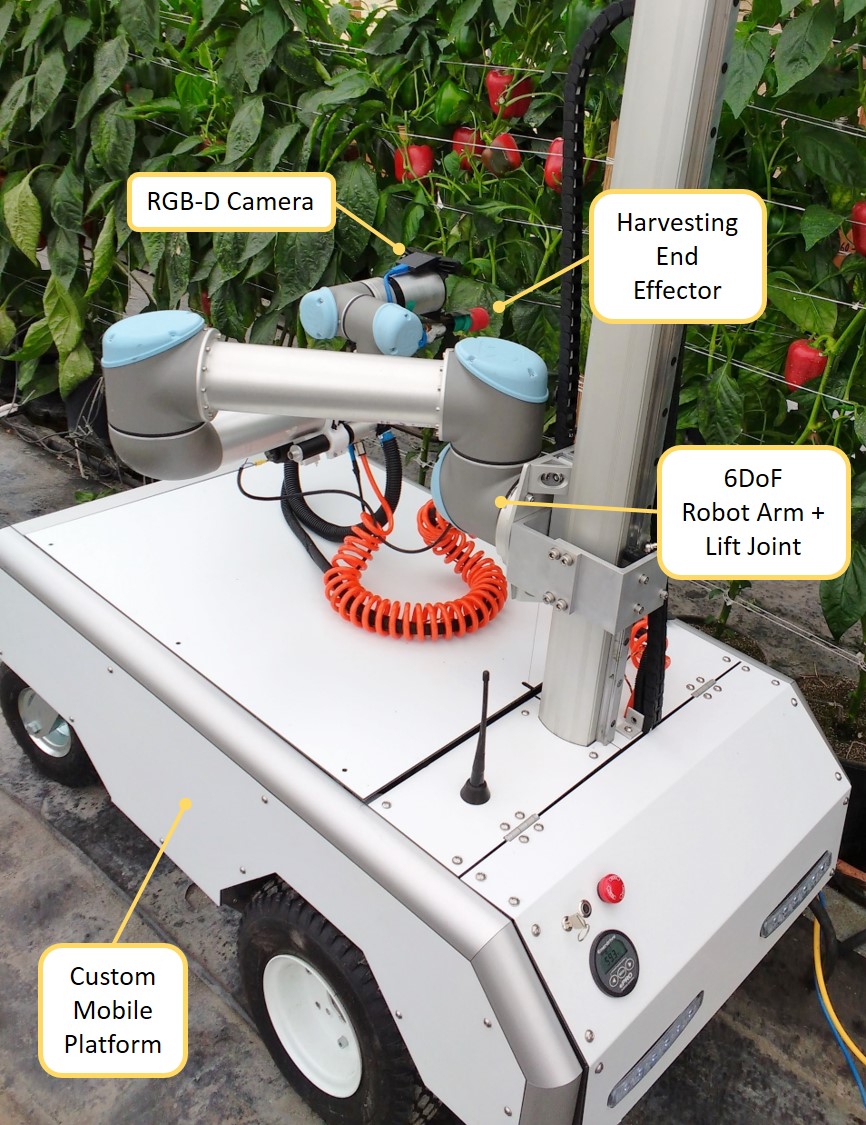}
 	\caption {Our mobile sweet pepper harvesting platform operating in a protected cropping environment. Harvesting is performed with a custom harvesting tool and 7DOF manipulator (6DOF articulated arm + lift joint) integrated into a custom differential drive mobile base.} 
 	\label{fig:frontpage1}
 	\label{fig:harvey_platform}
 	\vspace{-20pt}
 \end{figure}


In this paper, we present an autonomous sweet pepper harvester that works towards addressing these three key challenges.  We demonstrate a simple and effective vision-based algorithm for crop detection, a 3D localisation and grasp selection method, and a novel end-effector design for harvesting. To reduce complexity of motion planning and to minimise occlusions we focus on picking sweet peppers in a protected cropping environment (see Figure \ref{fig:harvey_platform}) where plants are grown on planar trellis structures. Experimental results with minor modifications to the plant (removal of some leaves) demonstrated a harvesting success rate of 58\%, a grasping success rate of 81\% and a detachment success rate of 90\% for favourable cultivar. We believe these results represent a significant improvement on the previous state-of-the-art and show encouraging progress towards the possibility of a commercially viable autonomous sweet pepper harvester. 


The remainder of the paper is structured as follows. A review of current state-of-the-art methods for autonomous harvesting of horticultural crops is presented in Section \ref{sec:literature}. The design of our autonomous harvesting platform is presented in Section \ref{sec:system_design} and methods for perception and planning are presented in Section \ref{sec:perception_and_planning}. Finally, results of two field experiments are presented in Section \ref{sec:results}, before a discussion of the key challenges and future work in Section \ref{sec:discussion_and_conclusion}.

\section{Literature}\label{sec:literature}


Autonomous harvesting has been demonstrated on a number of horticultural crops such as sweet peppers \cite{Hemming2014a,Bontsema2014}, cucumbers \cite{Henten2002}, citrus fruits \cite{Mehta2014}, strawberries \cite{Hayashi2010} and apples \cite{Bulanon2010a,De-An2011}. 
Autonomous sweet pepper harvesting has previously been demonstrated by the Clever Robots for Crops (CROPS) project \cite{Hemming2014a,Bontsema2014}. The CROPS platform achieved a harvesting success rate of 6\% for unmodified crop and 33\% when occluding leaves and crop clusters were removed. Their work highlights the difficulty and complexity of the harvesting problem. Several key research challenges remain before widespread commercial adoption can occur~\cite{Bac2014}. These challenges can be broken into three categories: perception, motion planning and hardware design.

\subsection{Perception}

Crop perception pipelines generally include detection, segmentation and 3D localisation stages. 

Detection and segmentation are essential in knowing the approximate location of the crop within an image and approaches have been developed for both automated harvesting~\cite{Kapach2012,Bac:2013aa,McCool:2016aa} and  yield estimation~\cite{Hung:2013aa}.





After detection and segmentation, the 3D location (position, orientation and shape) of the crop is determined using a 3D sensor such as a ToF camera~\cite{Hemming2014a,Lehnert2016}, stereo vision~\cite{Henten2002,Hayashi2010} or even a single-point laser range finder~\cite{Han2012}. This 3D localisation step allows a grasp and/or cutting pose to be determined.  Methods in the literature for 3D localisation are generally ad-hoc and crop specific. 




\subsection{Motion Planning}

Common methods of motion planning for autonomous crop harvesting include open loop planning \cite{Hemming2014a,Baur2014} and visual servoing \cite{Han2012,Hayashi2010}. Open loop planning methods sequentially sense, plan and act, leaving such systems susceptible to unperceived environment changes \cite{Barth2016}. Visual servoing methods require a high update rate, but are particularly useful for motion planning within dense vegetation where crop localisation from a single viewpoint can perform poorly due to occlusions~\cite{Barth2016}.


Reliable grasping in a dense and cluttered environment remains an active research topic \cite{Killpack2015}, often requiring tactile sensing to discern between rigid and deformable objects. As advocated in \cite{Bac2014}, simplifying the workspace or developing harvesting tools which simplify the harvesting operation can greatly improve the success of motion planning in cluttered horticultural environments.

\subsection{Harvesting Tools and Manipulators}

A range of manipulator configurations have been used for autonomous harvesting projects including 3DOF Cartesian, anthropomorphic arms and 6DOF manipulators \cite{Bac2014}. Several works have compared various joint configurations to optimise target-reachability in cluttered environments \cite{VanHenten2009, Lehnert2015}. 


A key component of any autonomous harvesting system is the harvesting end-effector that grasps and/or cuts the crop. Suction cups are a common gripping mechanism  \cite{Hemming2014a,Bontsema2014, Hayashi2010, VanHenten2003} that are mechanically simple and only require access to a single exposed face of the crop.  

Another gripping alternative are contact-based grippers \cite{De-An2011, Ling2005} which generally employ mechanical fingers that close around the crop. Contact-based grippers can grip the crop very securely, but are more prone to interference from nearby objects such as branches and other fruit. Some crops such as sweet pepper and cucumber must also be cut from the plant and so require an additional detachment tool such as a thermal cutter \cite{Henten2002} or scissor-like mechanism \cite{Hemming2014a,Hayashi2010,Han2012}. End effectors customised to a particular crop are common, for example in 2014 Hemming et al.~\cite{Hemming2014a} developed a custom harvesting tool which simultaneously envelops and cuts-free sweet peppers with a hinged jaw mechanism. This mechanism was found to be more effective than a scissor mechanism, however size and geometry constraints restricted access to some sweet peppers~\cite{Hemming2013b}.

\section{System Design}\label{sec:system_design} 
\begin{figure}[!b]
	\vspace{-15pt}
	\includegraphics[width=\columnwidth]{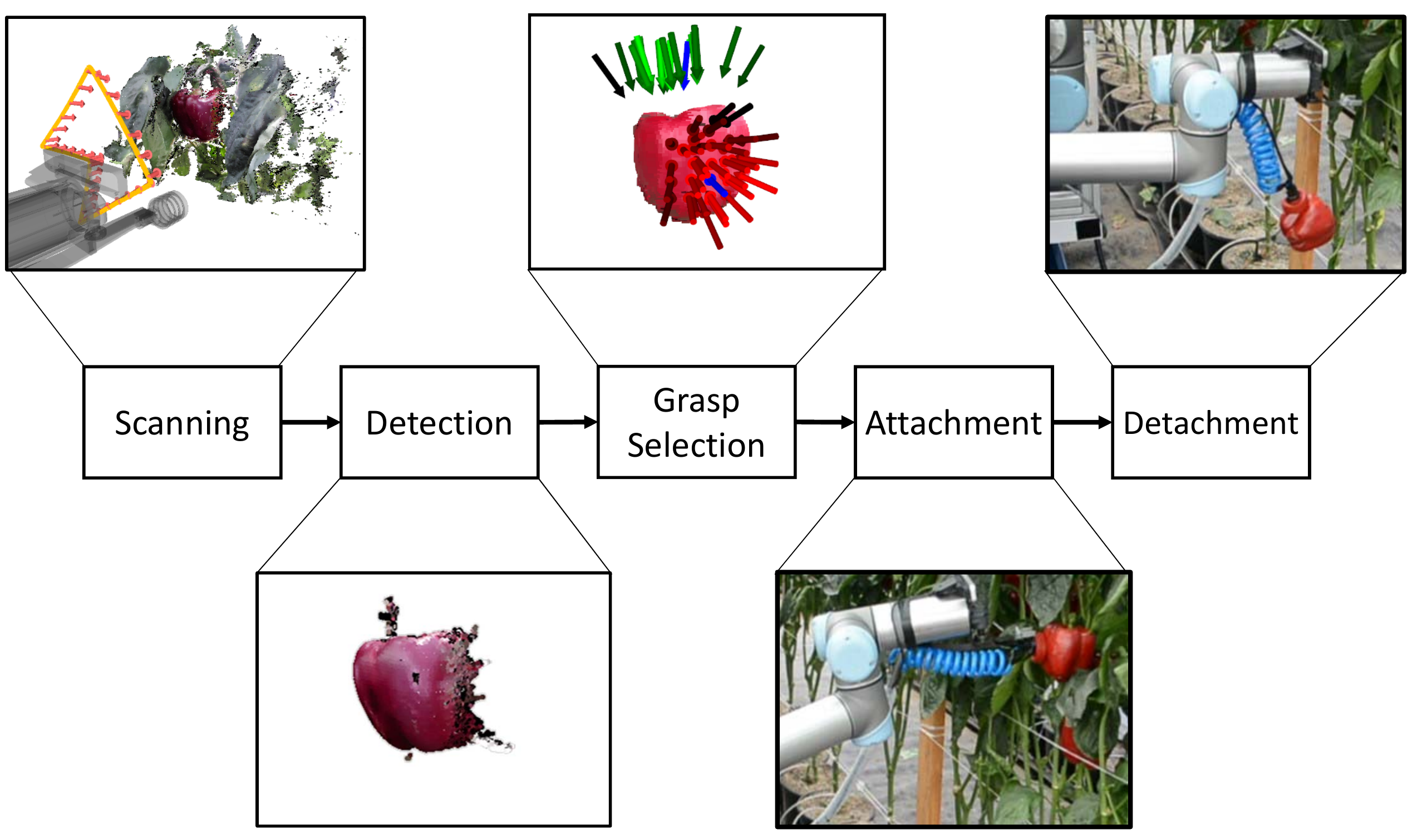}
	\centering
	\vspace{-10pt}
	\caption{The five stages of the autonomous harvesting cycle.} 
	\label{fig:system_stages}
\end{figure}

This section outlines the system design for our autonomous sweet pepper harvester. The overall procedure for harvesting sweet peppers is shown in Figure~\ref{fig:system_stages} and can be broken down into five key steps:
\begin{enumerate}
	\item \textbf{Scanning}: The robot arm is moved in a pre-determined scanning motion to build up a 3D model using an eye-in-hand RGB-D camera.
	\item \textbf{Crop Detection:} Sweet peppers from the 3D scene are segmented using colour information and localised by fitting a 3D parametric model to the segmented points.
	\item \textbf{Grasp Selection:} Candidate grasp poses are computed using the segmented sweet pepper point cloud.
	\item \textbf{Crop Attachment:} Suction cup grips the sweet pepper.
	\item \textbf{Crop Detachment}: Oscillating blade cuts the sweet pepper from the plant.   
\end{enumerate}

The first three steps (scanning, crop detection, and grasp selection) form part of the perception system described in Section \ref{sec:perception_and_planning}. 
The procedure for the crop attachment and crop detachment step, are described in Section \ref{sec:harvest_tool}.



\subsection{Protective Cropping Systems}\label{sec:cropping_environment}


Sweet peppers are grown in both fields and protected cropping environments. In this work, we focus on picking sweet pepper's in protected cropping environments. In these environments, sweet pepper's are grown on a two dimensional planar surface, significantly reducing occlusion and the need for complex collision avoidance and motion planning around the crop structure. Also, many protective cropping environments have a translucent outer surface that diffuses incoming sunlight; creating favourable conditions for computer vision with relatively even lighting. 

The protective cropping environment used in this work has crop grown in planar rows up to \unit[2]{m} tall with rows spaced approximately \unit[1]{m} apart. This layout informed the workspace requirement of our sweet-pepper harvesting platform design.

\subsection{Platform Design}

The harvesting robot ``Harvey'', is shown in Figure~\ref{fig:harvey_platform}. The custom differential drive platform was designed to work independently within a protected cropping environment and manoeuvre between crop rows for up to 8 hours powered by an internal 3kWh lead-acid battery. The platform has a 6DoF revolute arm (Universal Robotics UR5) mounted on a prismatic lift joint (Thomson LM80). The differential drive mobile base houses the batteries, drive motors, gearboxes, computer hardware, robot controller and forward facing laser scanner for mobile navigation and obstacle detection.


\begin{figure}[!b]
	\vspace{-15pt}
	\centering
	\includegraphics[width=\columnwidth]{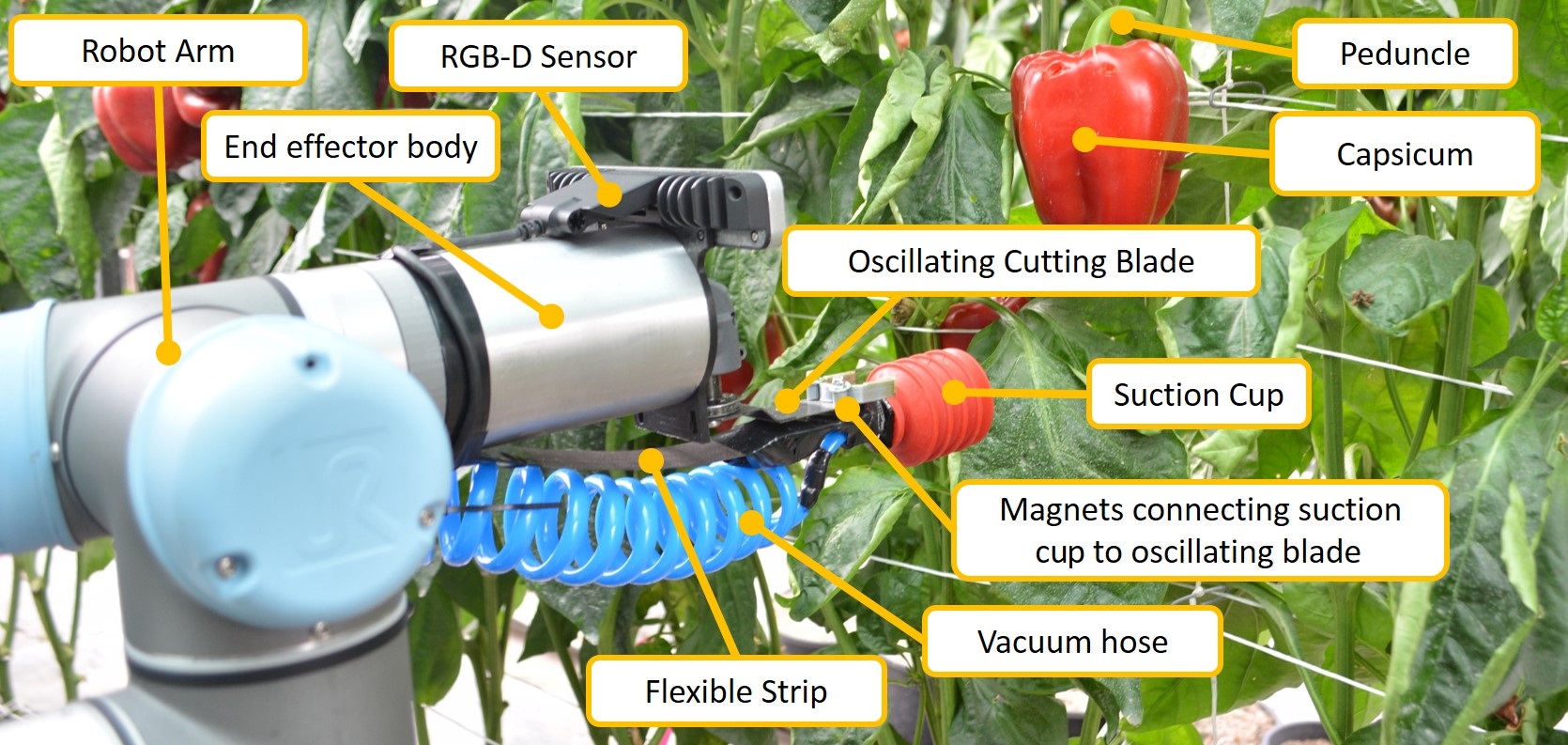}
	\centering
	\caption{Harvesting tool attached to the robot end effector}
	\label{fig:gripper_labelled}

\end{figure}

\subsection{Harvesting Tool Design} \label{sec:harvest_tool}

Our custom harvesting tool shown in Figure~\ref{fig:gripper_labelled} is able to grip sweet peppers with a suction cup, and then cut them free from the plant using an oscillating blade. 

Natural variation in crop size, shape and orientation make choosing a single end-effector pose to simultaneously grasp and cut each sweet pepper challenging and unreliable. To overcome this difficulty, a key feature of our harvesting tool design is a passive decoupling mechanism that allows the gripping and cutting operations to occur sequentially, at independently chosen locations. 

The decoupling mechanism is a flexible strip that tethers the suction cup to the body of the end effector. The suction cup is also magnetically attached to the underside of the cutting blade, allowing the robot arm to guide the suction cup during the attachment phase. After attachment, the cutting blade is lifted to decouple the suction cup from the cutting blade. The suction cup is then only attached to the end effector via the flexible tether, allowing the cutting blade to move independently of the suction cup through the cutting operation. After detachment, the sweet pepper falls from the plant and hangs freely from the flexible tether. The suction cup and cutting blade can be magnetically re-coupled ready for the next harvesting cycle using gravity by simply pointing the harvesting tool downwards. The sweet pepper is released into a collection crate by releasing the vacuum. 

This simple and passive decoupling method requires no additional actuators and allows for a greater harvesting success rate. 

%


The harvesting tool also contains an RGB-D camera (Intel\textregistered Realsense SR300 RGB-D) sensor for perceiving the crop and a micro-switch for checking whether the suction cup is coupled with the cutting blade. The body of the end effector contains a modified oscillating multi-tool for cutting stems. A pressure sensor on the vacuum line is able to detect sucessfull attachment of the suction cup.

\subsection{Software Design}\label{subsec:harvesting}
\begin{figure}[!tb]
	\centering
	\includegraphics[width=0.9\columnwidth]{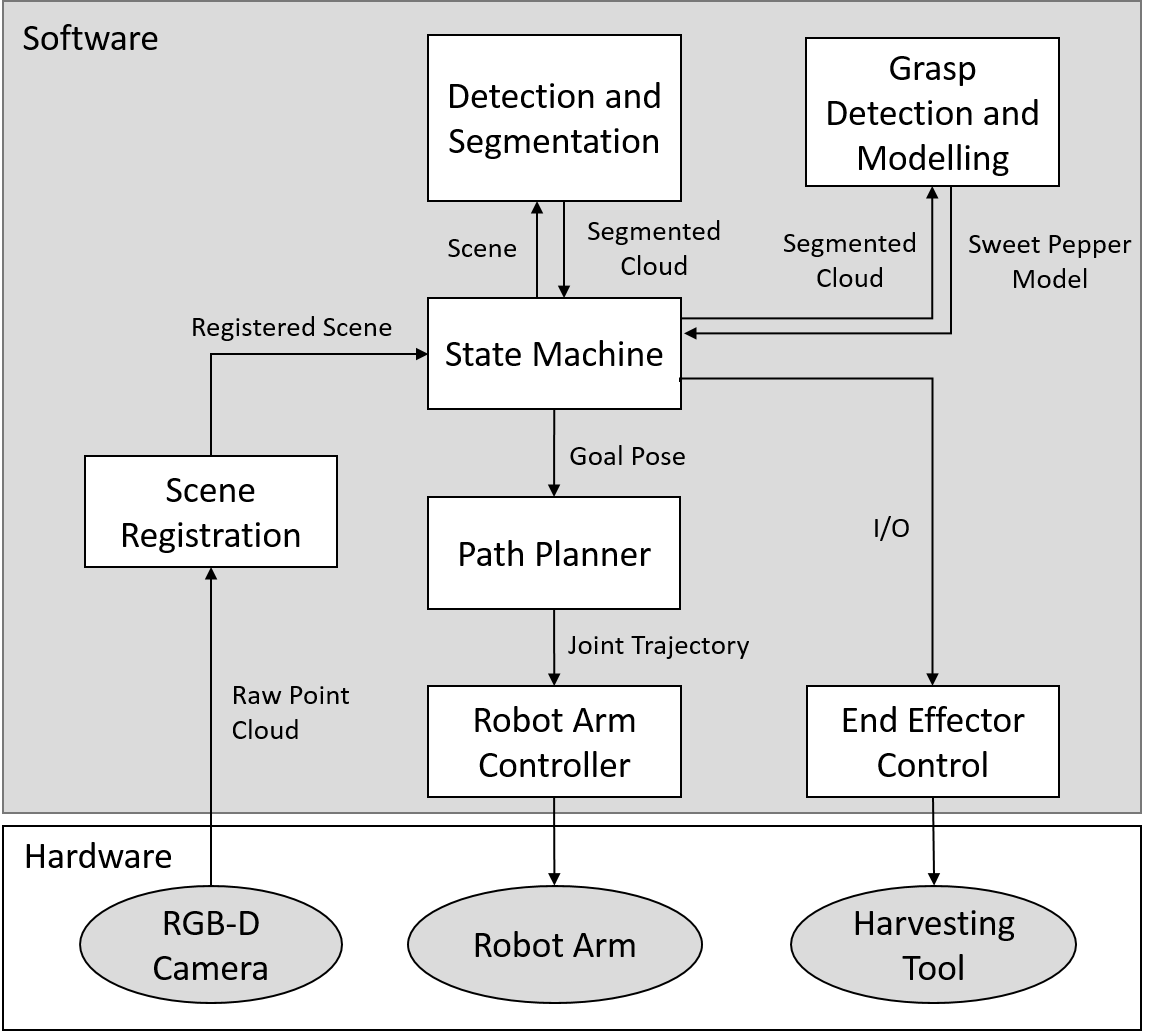}
	\caption{ System diagram illustrating how each subsystem is connected. Images from the RGB-D camera are fused by the scene registration node. The state machine uses the scene registration, detection and grasp detection subsystems to localise the sweet pepper and estimate grasp poses. Grasp poses are then used to perform the harvesting actions using the path planner, robot arm controller and end effector controller subsystems. }
	\label{fig:system_diagram}
	\vspace{-15pt}
\end{figure}

The software design uses the Robot Operating System (ROS) to communicate between independent processes (nodes).  Figure \ref{fig:system_diagram} illustrates the connection between software components.

The ROS MoveIt! \cite{moveit!} library was used for motion planning, with the TRAC-IK~\cite{Beeson2015} Inverse Kinematics (IK) solver for improved solution rate and solution time compared to standard IK solvers. Trajectory execution was performed by the Universal Robots ROS controller package. 

The Scene Registration node registers point clouds from the robot's RGB-D sensor into a smoothed colour point cloud using the Point Cloud Library (PCL) \cite{Rusu2011} implementation of the Kinect Fusion algorithm. This node does not use joint state information from the robot controller which avoids the need for accurate time synchronisation between RGB-D frames and joint states.

\begin{figure}[!b]
	\vspace{-20pt}
	\centering
	\includegraphics[width=\columnwidth]{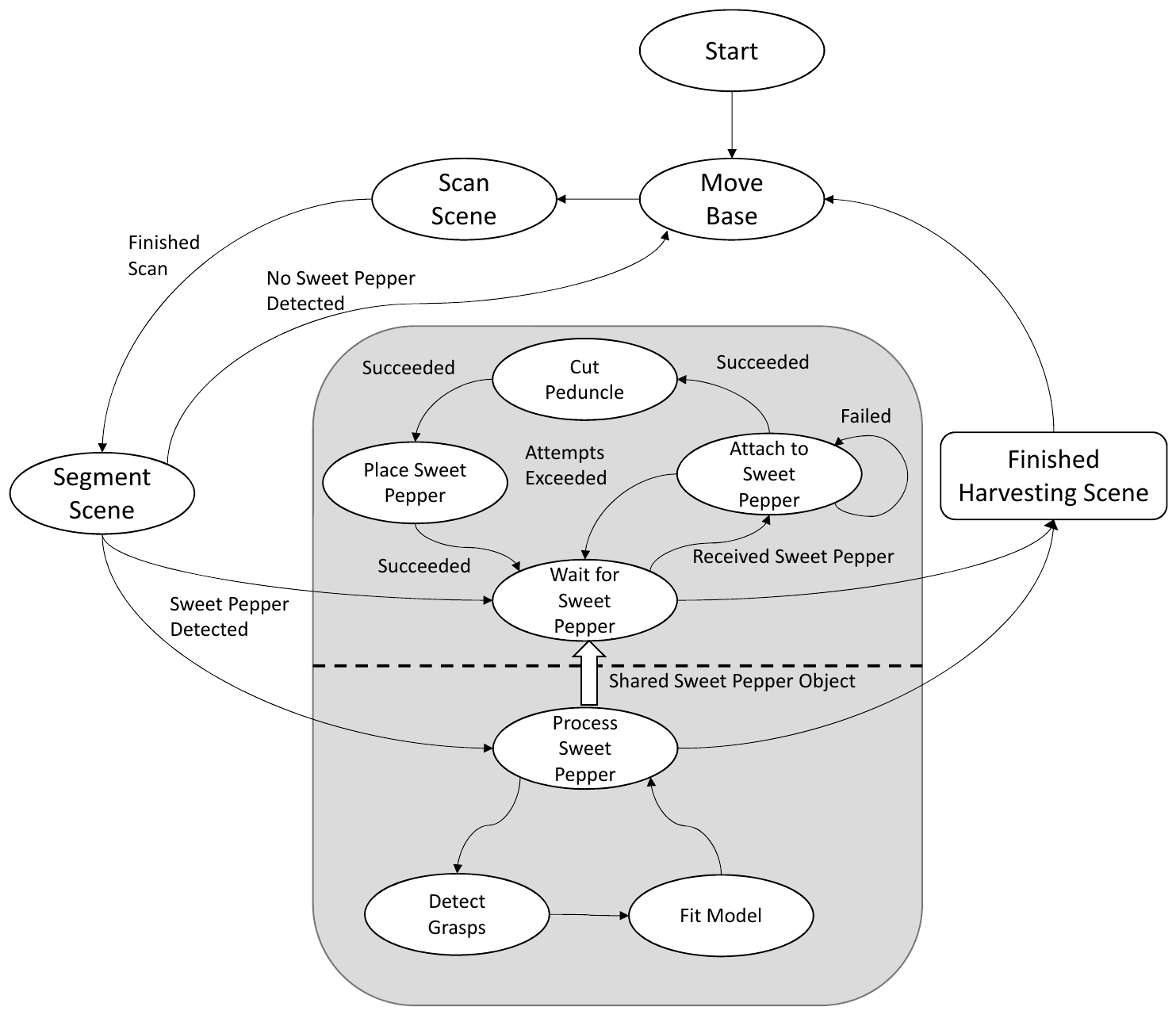}
	\vspace{-15pt}
	\caption{ State machine of the harvesting system illustrating the steps for harvesting sweet pepper. After scene segmentation completes, the sweet pepper processing states (below dotted line) run concurrently with harvesting states (above dotted line) reducing the overall harvesting time. }
	\label{fig:state_machine}
\end{figure}

The Detection and Segmentation node detects sweet peppers within a scanned scene. The node returns a list of point cloud segments and is implemented as a ROS service client. 

The State Machine node implements the harvesting logic and coordinates the operation of other nodes to perform the harvesting operation. Figure \ref{fig:state_machine} outlines the logic implemented in this state machine for a single harvesting cycle.

Following segmentation of the scene into individual sweet peppers, two concurrent state machines are spawned. 
The first state machine (below the dotted line in Figure \ref{fig:state_machine}) continuously processes segmented sweet peppers to calculate candidate grasping and cutting poses. The second state machine~(above the dotted line in Figure \ref{fig:state_machine}) waits for a processed sweet pepper, then plans and executes the harvesting operation. This concurrent processing and harvesting of sweet peppers speeds up the harvesting process when a scene contains multiple detected sweet peppers. 

Failure to attach the suction is detected via an in-line pressure sensor on the vacuum system and triggers a re-attachment at the next-best grasp candidate. After a fixed number of failed attachment attempts the state machine moves on to harvesting the next sweet pepper. Future work will also look into methods of sensing successful peduncle cutting to enable the state machine to attempt multiple cutting poses.



\section{Perception and Planning}\label{sec:perception_and_planning}
\newcommand{\vecx}{\mathbf{x}}
\newcommand{\mean}{\boldsymbol{\mu}}
\newcommand{\variance}{\boldsymbol{\Sigma}}

In this section we present our perception system for detecting, segmenting, and estimating the pose of sweet peppers, followed by the planning steps to perform the picking action. 

This firstly involves moving the RGB-D camera in a scanning motion, combining these views into a single 3D model, and segmenting sweet peppers using colour information as described in Section ~\ref{subsec:scanning_detection}. 
Grasp poses are then calculated by one of two methods described in Section \ref{subsec:grasp_selection}.
Grasp poses are then used by the subsequent motion planning system (Section \ref{subsec:motion_planning}) to perform the harvesting operation.

An overview of the perception system is given below where a more in depth description of the perception system can be found in our previous work \cite{Lehnert2016}. 

\subsection{Scanning and Sweet Pepper Detection}
\label{subsec:scanning_detection}

The initial stage of the perception process involves scanning the crop row with the end-effector mounted RGB-D camera. 
The point clouds captured from the RGB-D sensor during the scanning motion are used as input to the point cloud registration method that creates a single 3D model of the scene using the Kinect Fusion algorithm. The RGB-D camera operates at approximately 30 frames per second with a depth resolution of \unit[2]{mm} and range between \unit[0.2]{m} to \unit[1.5]{m}.

Combining multiple views in a single 3D model helps to reduce point cloud noise, and reduces the effect of occlusion by leaves since sweet peppers will generally be seen clearly from at least one view during the scanning process. 

\begin{figure*}[!t]
	\centering
	\begin{subfigure}[t]{0.49\textwidth}
		\centering
		\includegraphics[width=0.85\columnwidth]{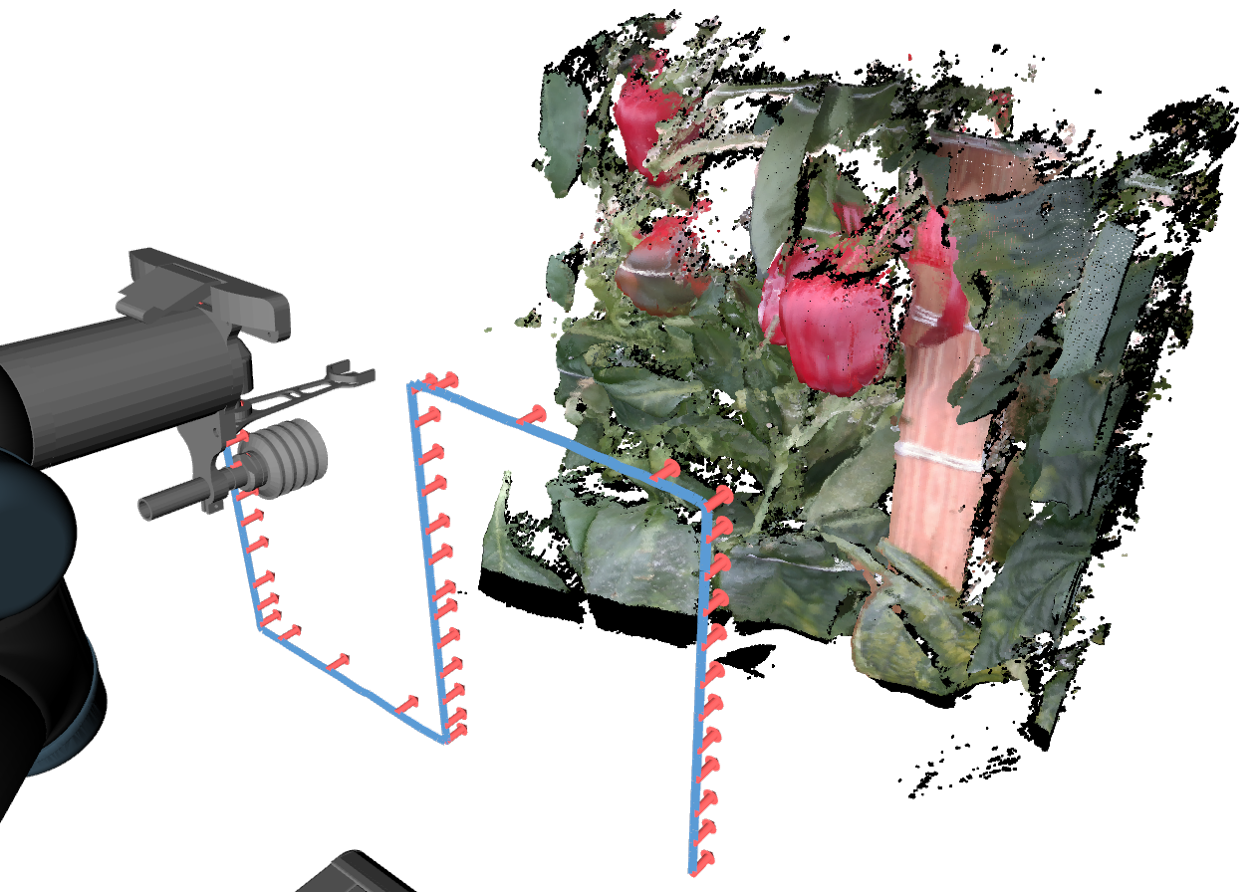}	
		\caption{}
	\end{subfigure}
	\begin{subfigure}[t]{0.49\textwidth}
		\centering
		\includegraphics[width=0.7\columnwidth]{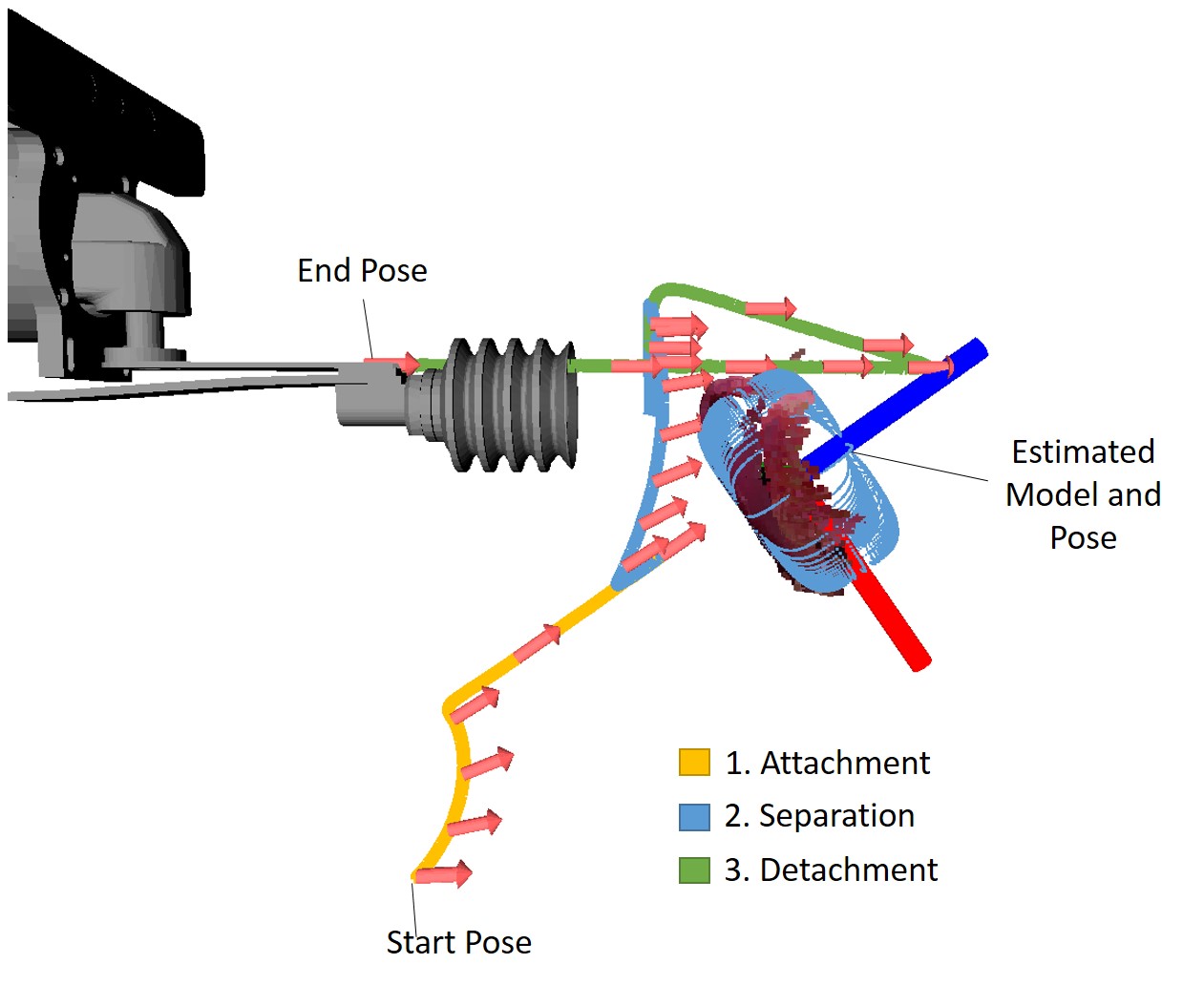}
		\caption{}
	\end{subfigure}
	\vspace{-5pt}
	\caption{(a) Scanning trajectory of the camera during a single field trial along with the registered point cloud (b) End effector trajectory of a single harvesting trial. The end effector position and orientation is indicated by a coloured line and red arrow, respectively. The trajectory begins (start pose) with the attachment stage (yellow line), transitioning into the separation stage (blue line) and finishing with the detachment stage (green line). The estimated pose of the sweet pepper is shown using blue and red axes. }
	\label{fig:scanning}
	\label{fig:harvest_traj}
	\vspace{-20pt}
\end{figure*}

Two types of scanning trajectories have been used, a diamond and boustrophedon (lawnmower) pattern with a fixed offset between the camera and crop row. Figure \ref{fig:scanning} shows the trajectory (blue line) and pose (red arrows) of the RGB-D camera on the end effector recorded while scanning a sweet pepper as part of the experiment presented in Section \ref{subsec:experiment}. The figure also shows the resulting registered point cloud generated from the Kinect Fusion algorithm produced during the scan. Scanning is currently performed perpendicular to the crop row, as this was simple to define and sufficient to build a detailed 3D model. This scanning method is inefficient in time and future work could look at selecting the next best view to maximise 3D information about a target crop.  

\newcommand *\naive {Na\"{i}ve }

The second stage of the perception system segments the sweet pepper from the background (leaves and stems). This task is challenging due to variation in crop colour and illumination as well as high levels of occlusion. Prior work by McCool et al.~\cite{McCool:2016aa} proposed a highly accurate sweet pepper segmentation approach capable of segmenting both green and red sweet pepper. In this work, we are interested in harvesting red sweet pepper only and so use a simpler and faster colour-based segmentation approach outlined in \cite{Lehnert2016} which applies a \naive Bayes classifier to points in the fused point cloud in a rotated-HSV colour space. Parameters of the HSV model were determined by computing the mean and variance on a sample of sweet peppers in the field.


Once segmented, a Euclidean clustering step is used to group the remaining points into multiple distinct sweet peppers. Clustering is based on a minimum distance threshold, selected as \unit[2]{mm} corresponding to the resolution of the depth camera. A limitation of this method is when two sweet peppers are in contact with each other. If multiple sweet peppers are in view this clustering step also determines the best candidate sweet pepper based on the cluster which has the largest number of detected 3D points (most information available) and has the closest centroid to the end effector. Lastly, smoothing and outlier removal is performed on the points to filter noise from the detection step.

\subsection{Grasp Selection}
\label{subsec:grasp_selection}
Grasp poses for each sweet pepper are calculated using the the segmented 3D point cloud of a sweet pepper.
An ideal grasp pose will place the suction cup squarely on a planar region of a sweet pepper. In this work we developed two different methods for selecting both grasp and cutting poses.

The first grasp selection method is outlined in  \cite{Lehnert2016} and fits a geometric model (superellipsoid) to the point cloud to estimate the sweet peppers size, position and orientation. The grasp pose is calculated to be in the centre of the front face of the sweet pepper, while the cutting pose is calculated to be offset from the top face of the sweet pepper.

The second method takes a different approach and selects multiple grasp poses directly from the point cloud data by calculating surface normals with a fixed patch size. These surface normals can be used directly as grasp poses however, we rotate these poses along their axis to keep the end effector upright, minimising large jumps in the wrist configuration. 

Candidate poses are ranked based on a utility function that is the weighted average of three normalised scores $S_{i1}, S_{i2} \text{ and } S_{i3}$ based on the surface curvature, distance to the point cloud boundary and angle with respect to the horizontal world axis, respectively, where $i$ is the current candidate pose. This utility function favours grasp poses that are close to the centre of the sweet pepper, on planar surfaces, aligned with the horizontal world axis and away from discontinuities caused by occlusion. 
The \textit{utility}, $U_i$ of the grasp pose $i$ is calculated according to
\begin{equation}
U_i=\sum_{j=1}^{3} W_j \, S_{ij}, \text{ given } \sum^{3}_{j=1} W_j  =1
\end{equation}
where $0 \le U_i \le 1$, $0 \le S_{ij} \le 1$ is the \textit{score} of the grasp pose $i$ and $W_j$ are \textit{weighting coefficients} that describe the importance of each score.

An advantage of this method is it finds multiple grasp poses, in contrast to a single grasp pose when using the model fitting method.

Cutting poses for this second method are estimated in a similar way to grasp poses, however with a modified utility function that favours vertical rather than horizontal normal vectors. This tends to find normal vectors on the top surface of the sweet pepper in a ring around the peduncle. We reject poses with low utility to remove outliers, translate them vertically to avoid damaging the sweet pepper and create the final cutting pose using the medians of the x, y and z-axes independently. The orientation of the final cutting pose is set so as to keep the end effector level and perpendicular to the plane of the crop row.

Both our strategies for calculating the cutting pose (parametric model fitting and surface normal estimation) assumes the sweet pepper is vertical with a peduncle near the top. We are currently investigating methods of directly detecting the peduncle location in the colour point cloud to accommodate highly angled sweet peppers. 

\begin{figure}[!t]
	\centering

	\begin{subfigure}[b]{\columnwidth}\centering
		\includegraphics[width=0.3\columnwidth]{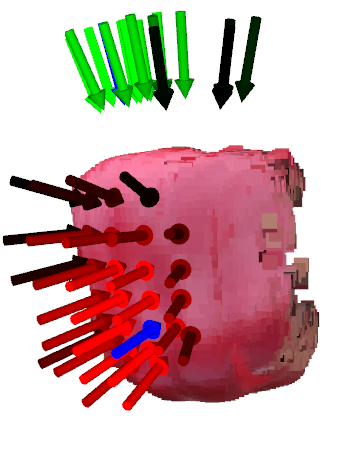}
		\includegraphics[width=0.35\columnwidth]{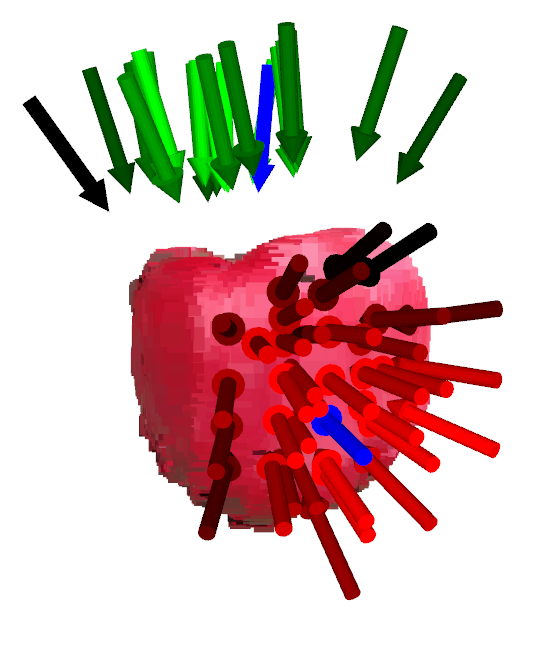}
	\end{subfigure}
	\vspace{-15pt}
	\caption{Computed grasp poses. The utility of each grasp pose is shown as the gradient from red to black, where black is the lowest utility. The blue arrow indicates the best pose. The vertical green arrows are estimated cutting poses for potential peduncle locations}
	\label{fig:grasp_detection}
	\vspace{-15pt}
\end{figure}

An example of the grasp selection method applied to real sweet pepper point clouds is shown in Figure \ref{fig:grasp_detection} where the utility of the grasping and cutting poses (surface normals) are represented as the gradient from red or green to black, where black has the lowest utility. 

\subsection{Motion Planning} \label{subsec:motion_planning}
Harvesting trajectories are calculated relative to the grasping and cutting poses described in the previous section. The attachment trajectory starts at a fixed offset back from the grasping pose and moves the suction cup along the selected approach axis. This motion, as computed by the planner, can be seen as the yellow line within Figure \ref{fig:harvest_traj}.

Once the attachment trajectory has been executed to attach the suction cup, the end effector is moved vertically from the attachment pose in order to decouple the suction cup from the cutting tool (as described in Section \ref{sec:harvest_tool}). 

Finally, a cutting trajectory is computed. This cutting trajectory is calculated so as to keep the end effector aligned with the horizontal world frame, as this was found to outperform trajectories aligned with the estimated orientation of the sweet pepper. 


The resulting end effector trajectory for the attachment, separation and detachment stages for one harvesting cycle is shown in Figure \ref{fig:harvest_traj}. 


As discussed in Section~\ref{sec:cropping_environment}, operating in a protected cropping environment results in a relatively planar workspace which simplifies the motion planning task for the robot. In this work, we avoid implementing complex obstacle avoidance around branches and leaves since in nearly all cases the robot performs ``in and out'' motions thanks to novel harvesting tool design. The motion planner of the robot takes into account self collisions with the robot arm and base platform, as well as a simple planar collision obstacle placed slightly behind the target crop row.

\section{Experiment and Results}\label{subsec:experiment}

\begin{table}[!b]
	
	\vspace{-8pt}
	\centering
	\caption{Scanning Parameters}
	
	\begin{tabulary}{\columnwidth}{lp{5cm}c}
		\toprule
		Trial & Parameter                               &              Value               \\ \midrule
		1     & Boustrophedon (Width, Height, Segments) & \unit[0.35]{m}, \unit[0.4]{m}, 3 \\
		      & Scan row offset                         &          \unit[0.25]{m}          \\
		      & Scanning speed                          &         \unit[0.1]{m/s}          \\ \midrule
		2     & Diamond (Radius)                        &          \unit[0.4]{m}           \\
		      & Scan row offset                         &          \unit[0.3]{m}           \\
		      & Scanning speed                          &         \unit[0.1]{m/s}          \\ \bottomrule
	\end{tabulary}
	\label{tab:scanning_parameters}
	
\end{table}

Two field trials were conducted over two harvesting seasons on a farm in North Queensland (Australia) within a protected cropping system. Overall the robot platform has been tested on a total of 75 sweet peppers in a real protected cropping system. Within this work, two different sweet pepper cultivars were trialled; the Claire cultivar in the first field trial and the Redjet cultivar in the second.

Trial 1 applied the lab tested grasping methods presented in \cite{Lehnert2016} in a real protected cropping environment. Changes to the platform between field trial 1 and 2 were also made in order to improve the autonomous behaviour of the system. These differences include:
\begin{itemize}
	\item The scanning trajectory was changed from a diamond pattern to boustrophedon pattern
	\item an upgrade from a manual scissor lift base platform (see \cite{Lehnert2016}) to a custom mobile platform (Figure \ref{fig:frontpage1}),
	\item the addition of a prismatic lift joint (integrated into the motion planner) replacing the manual lift joint to improve the workspace and reduce planning failures,
	\item the addition of a vacuum sensor to detect successful attachment and a micro-switch to detect decoupling of the suction cup and cutting blade.
\end{itemize}

\subsection{Methodology}

A formal experiment was conducted on the final day of each field trial, involving 24 and 26 sweet peppers for the first and second trial, respectively. Each trial was conducted within a single \unit[10]{m} stretch of a crop row. During the first field trial initial testing resulted in successfully picking a total of 25 additional sweet peppers, though not under strict experimental conditions.  

The methodology for each experiment is as follows. The base platform was placed at the start of a crop row and manually moved down the crop row after attempting to harvest a set of sweet peppers. The platform was positioned in order for the robot arms workspace to be within range of the next set of sweet peppers. For the first field trial, this involved physically pushing the platform and manually operating the scissor lift joint. For the second trial, the robot was remotely driven down the crop row, simulating an autonomous move base state (see Figure \ref{fig:state_machine}). There was no need to position the lift joint manually during the second trial as this was handled by the motion planner. 
 
If a failure occurred during an attempt, the robot arm moved back to its start position and the attempt was retried. If obstructions or occlusions caused multiple failed attempts, the scene was modified by either removing leaves or by adjusting the position of the sweet pepper, and then re-attempted.

\begin{table}[!b]
	
	\vspace{-8pt}
	\centering
	\caption{Parameters used for ranking grasp poses.}
	\label{tab:grasp_parameters}
	\begin{tabular}{lc}
		\toprule
		Parameter                                   &     Value     \\ \midrule
		Grasp utility weights ($W_1$, $W_2$, $W_3$) & 0.2, 0.5, 0.3 \\
		Surface Normal Patch Size                   &    0.025 m    \\ \bottomrule
	\end{tabular}
	
	\vspace{-5pt}
\end{table}

During each attempt, an attachment (the suction cup attached to the sweet pepper) and detachment (the peduncle was cut) success or failure was recorded. Additional notes were also recorded through the experiment and categorised as: major or minor sweet pepper damage (XD or MD), whether the sweet pepper was visually occluded or physically obstructed (OC or OB), or if the sweet pepper had an irregular shape (IS).

The scanning parameters such as size, speed and row offset for each trial are given in Table \ref{tab:scanning_parameters}. These parameters were empirically determined based on the protected cropping system and sweet pepper cultivar used. The scanning trajectories chosen did not change within the respective field trial.

The parameters for ranking surface based grasps are also given in Table~\ref{tab:grasp_parameters}.

\subsection{Results}\label{sec:results}

The success rates for the both field trials are presented in Table \ref{tab:harvest_result}. More in depth results for each of the 24 trials during the final experiment are shown in Table \ref{tab:individual_harvest_result}. A video of the robotic harvester demonstrating these results is available at \url{https://youtu.be/Fm70wHJ_Lew}.

Out of the total sweet peppers, $14/24$ (58\%) and $11/26$ (42\%) were successfully harvested for field trial 1 and 2 respectively, where a successful harvest includes a successful attachment and detachment. During field trial 1 and 2, a total of $22/24$ (92\%) and $11/26$ (42\%) sweet peppers were successfully detached from the plant, irrespective of an attachment success. On the other hand, attachment rates of $14/24$ (58\%) and $21/26$ (81\%) were achieved for field trial 1 and 2, respectively.

\newcolumntype{s}{>{\centering\hsize=.5\hsize\arraybackslash}X}
\newcolumntype{Y}{>{\raggedright\hsize=.5\hsize\arraybackslash}X} 

\begin{table}[!b]
	\vspace{-10pt}
	\centering
	\caption{Harvesting Result}
	\begin{tabularx}{\columnwidth}{ls|s}
		\toprule
		                 & Trial 1 - Cultivar: Claire & Trial 2 - Cultivar: Redjet   \\ \midrule
		Detach Rate      & \textbf{92}\% ($22/24$)    & 42\% ($11/26$)               \\
		Attach Rate      & 58\% ($14/24$)             & \textbf{81}\% ($21/26$)      \\
		Attach \& Detach & \textbf{58}\% ($14/24$)    & 42\% ($11/26$)               \\ \bottomrule
	\end{tabularx}%
	\label{tab:harvest_result}%

\end{table}%

\begin{figure}[!t]
	\centering
	\begin{subfigure}[b]{0.3\columnwidth}\centering
		\includegraphics[width=\columnwidth]{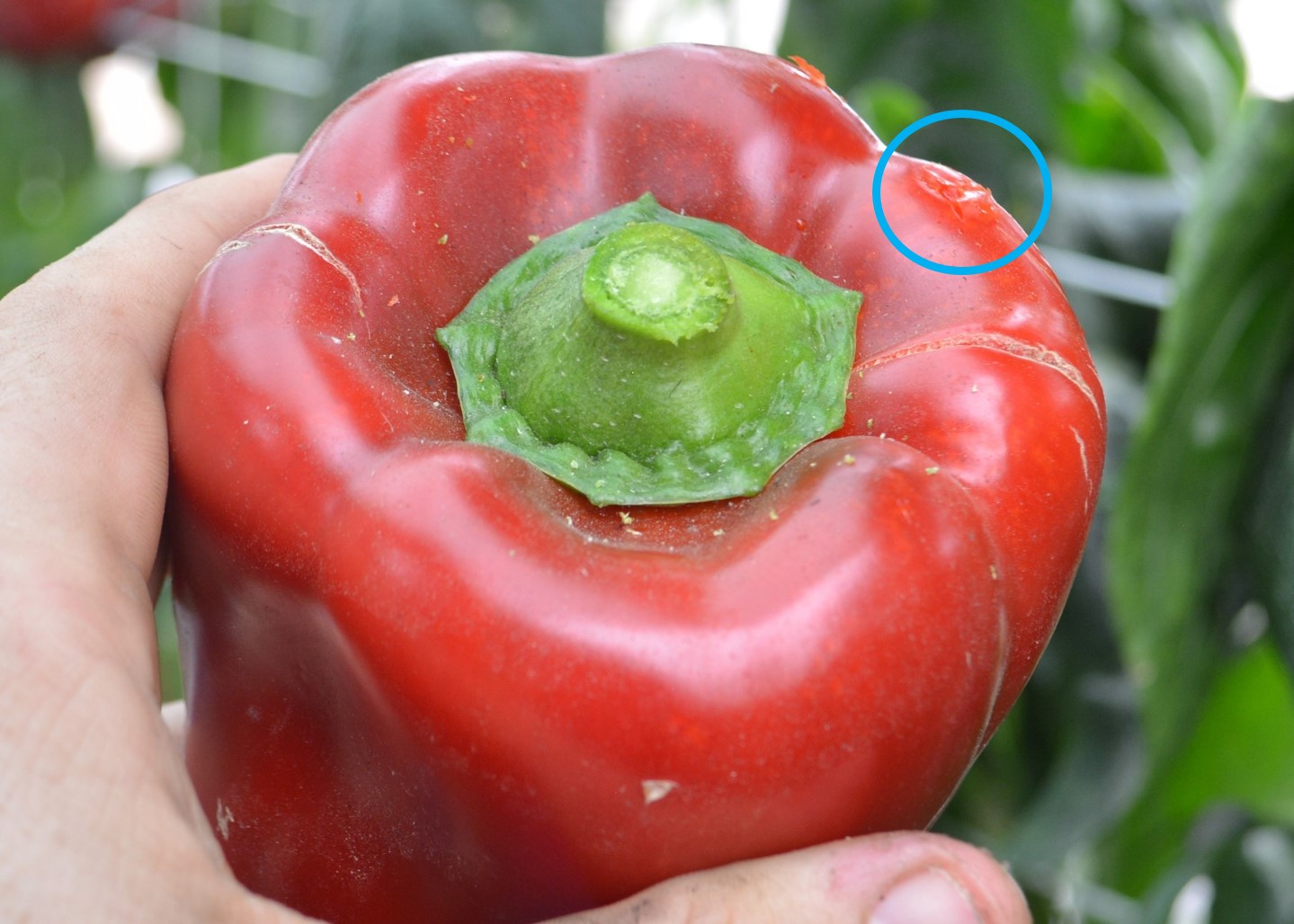}
		\caption{ }
		\label{fig:example_sweet_peppers:a}
	\end{subfigure}
	\begin{subfigure}[b]{0.3\columnwidth}\centering
		\includegraphics[width=\columnwidth]{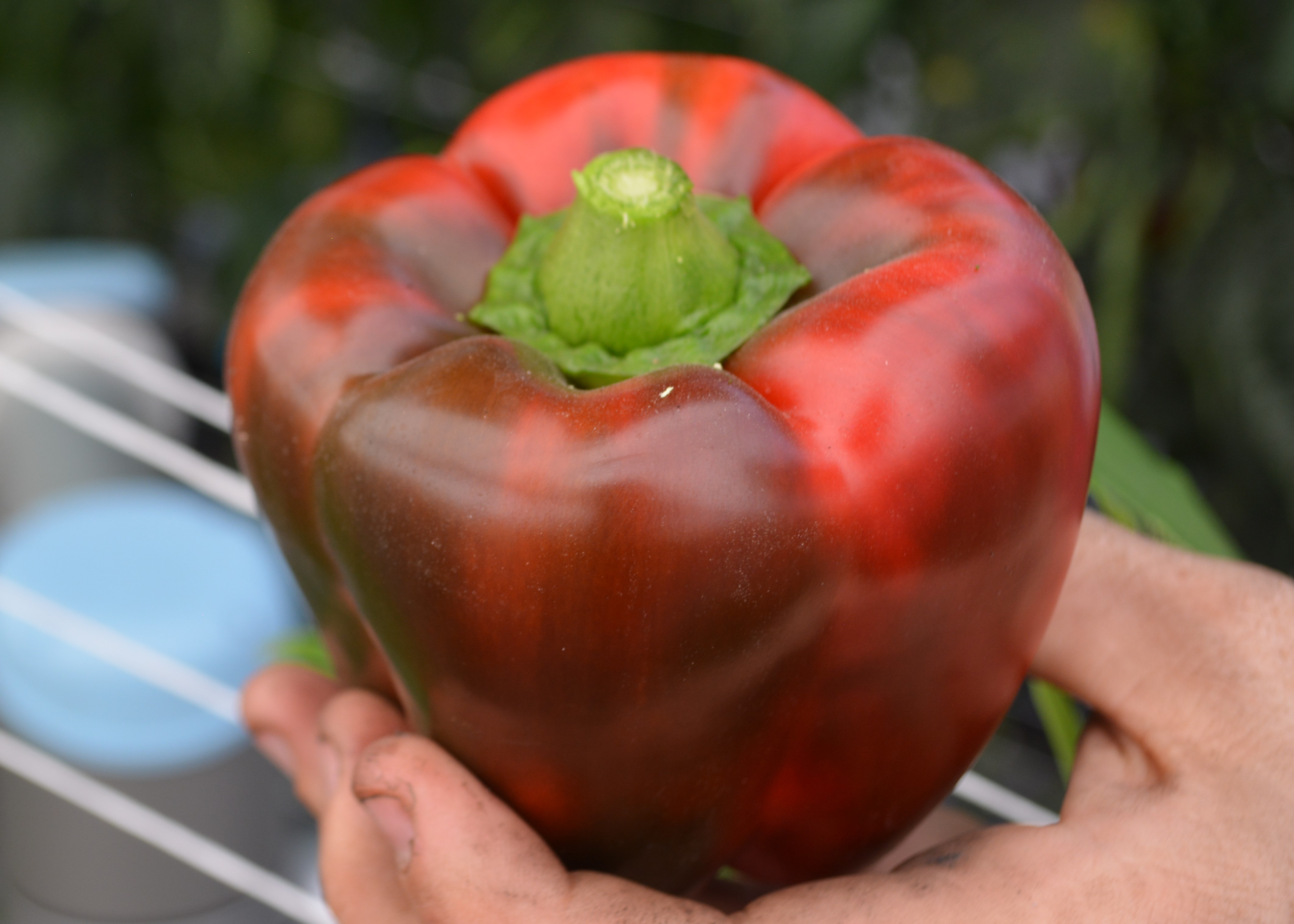}
		\caption{ }
		\label{fig:example_sweet_peppers:b}
	\end{subfigure}
	\begin{subfigure}[b]{0.3\columnwidth}\centering
		\includegraphics[width=\columnwidth]{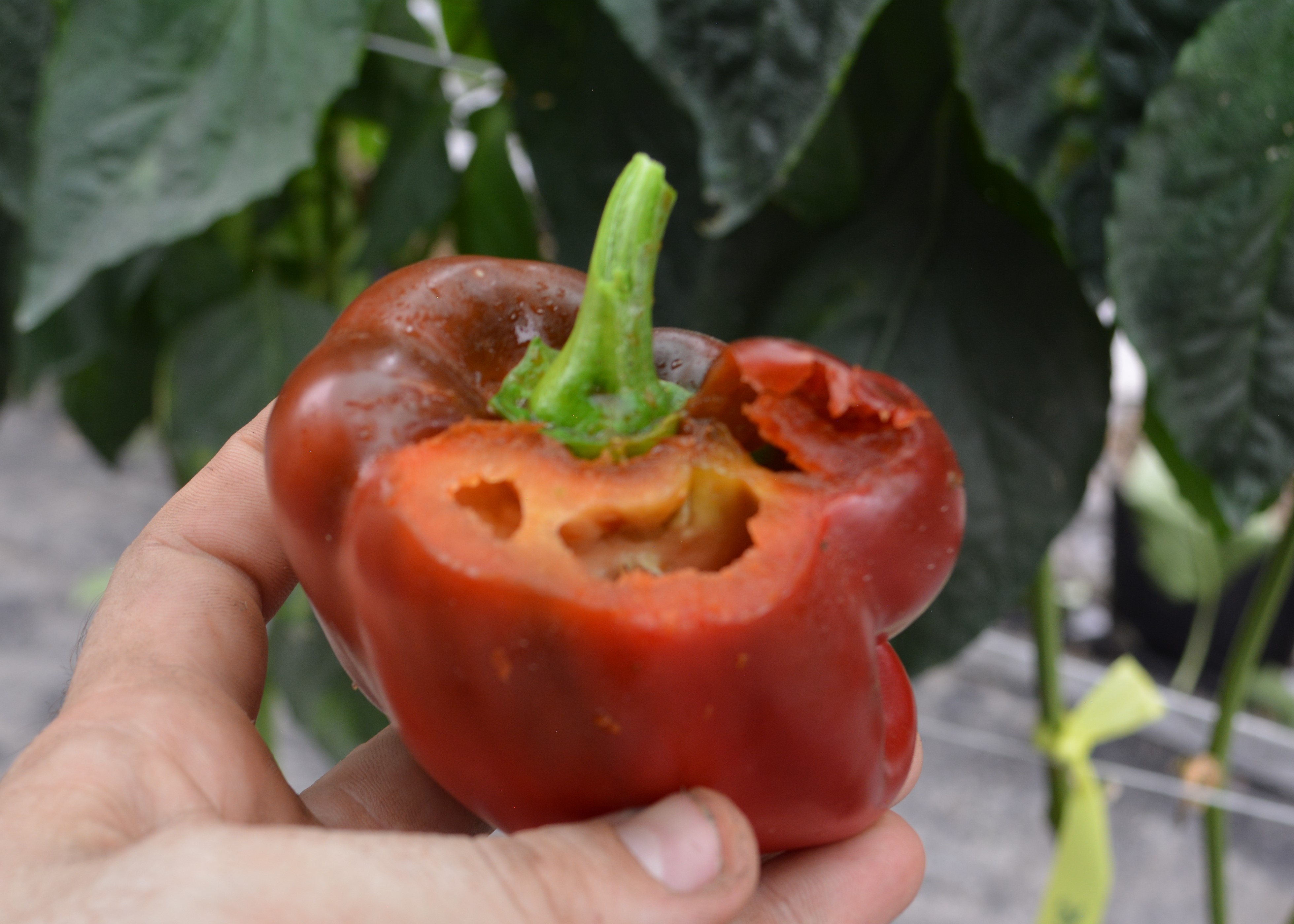}
		\caption{ }
		\label{fig:example_sweet_peppers:c}
	\end{subfigure}	
	\vspace{-5pt}
	\caption{(a) Harvested sweet pepper with minor damage circled in blue. (b) Example harvested sweet pepper. (c) Failed harvesting of a sweet pepper with major damage to its top face. It can be seen that this sweet pepper has large irregularities causing a poor estimate of its pose. }
	\label{fig:example_sweet_peppers}
	\vspace{-15pt}
\end{figure}

An example of two cases where detachment was unsuccessful (sweet peppers 6 and 19 within field trial 1) the sweet pepper was irregularly shaped which resulted in a poor grasp pose estimation. In the first case (sweet pepper no. 6), this led to the cutting blade repeatedly missing the peduncle, whereas in the second case (sweet pepper no. 19), the cutting blade caused major damage to the sweet pepper (see Figure \ref{fig:example_sweet_peppers:c}).

The average picking time was 35-40 seconds which includes the total time to perform a scan of each sweet pepper (approximately 15 seconds), fitting the model, pose and motion planning (5-10 seconds) and execution of the plan on the robot arm (10-15 seconds). The average number of attempts for each sweet pepper was 2.0 and 2.2 for field trial 1 and 2 respectively. 

\begin{figure}[!t]
	
	\begin{subfigure}[b]{\columnwidth}\centering
		\includegraphics[width=0.32\columnwidth]{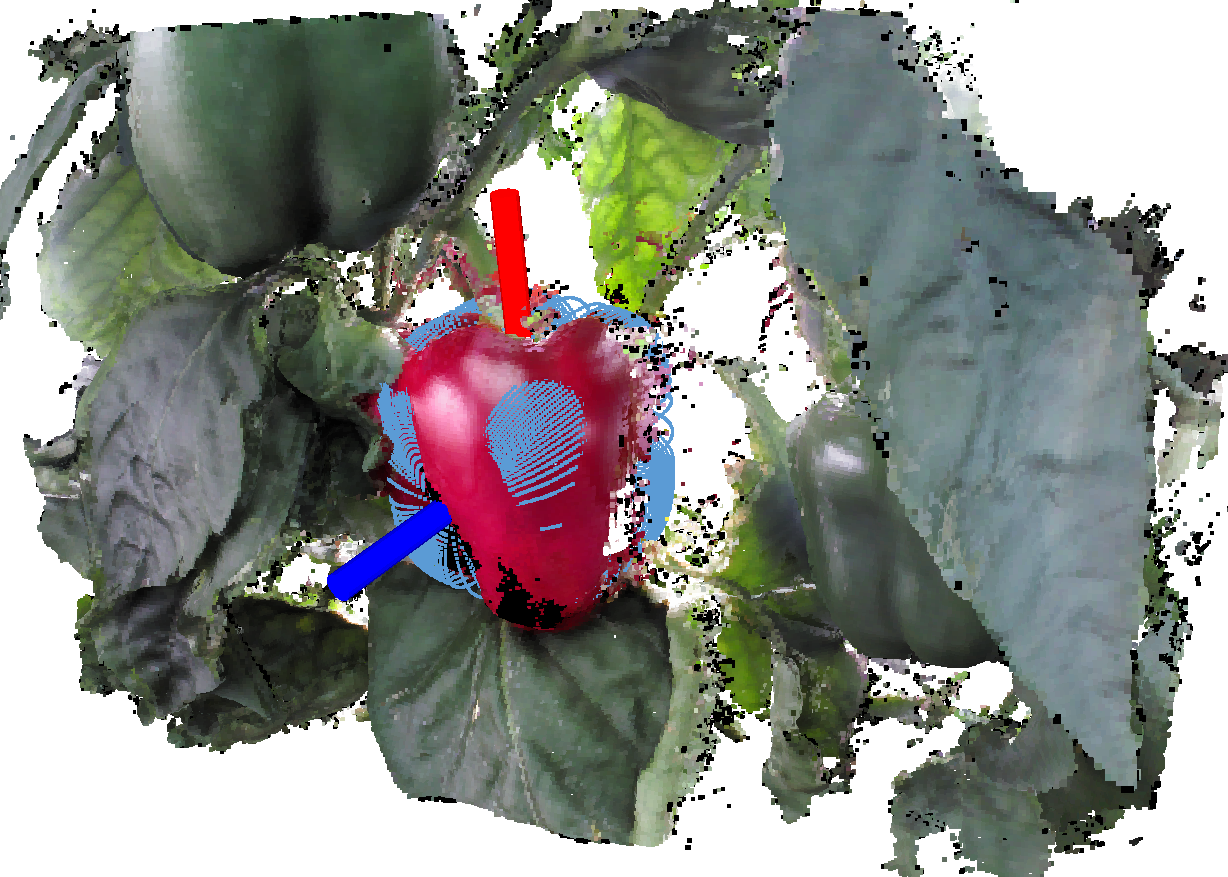}
		\includegraphics[width=0.32\columnwidth]{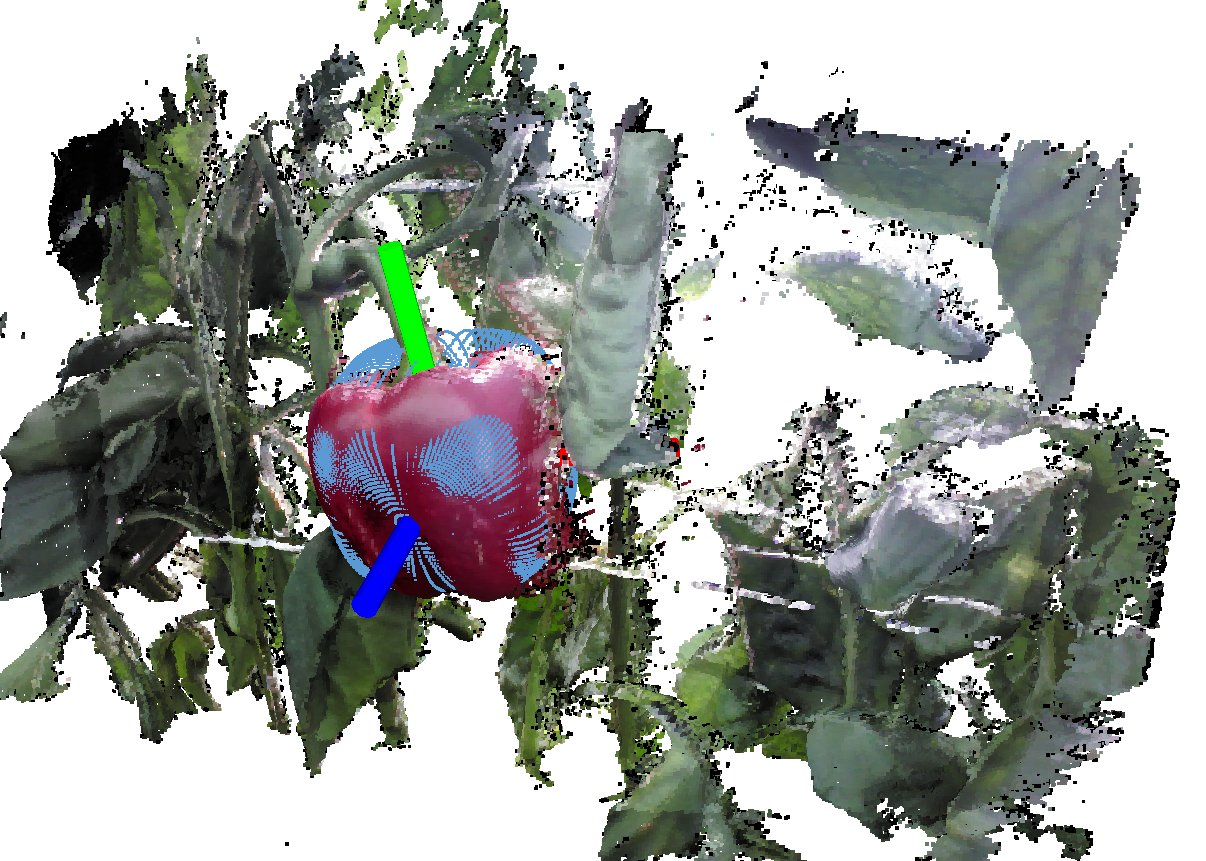}
		\includegraphics[width=0.32\columnwidth]{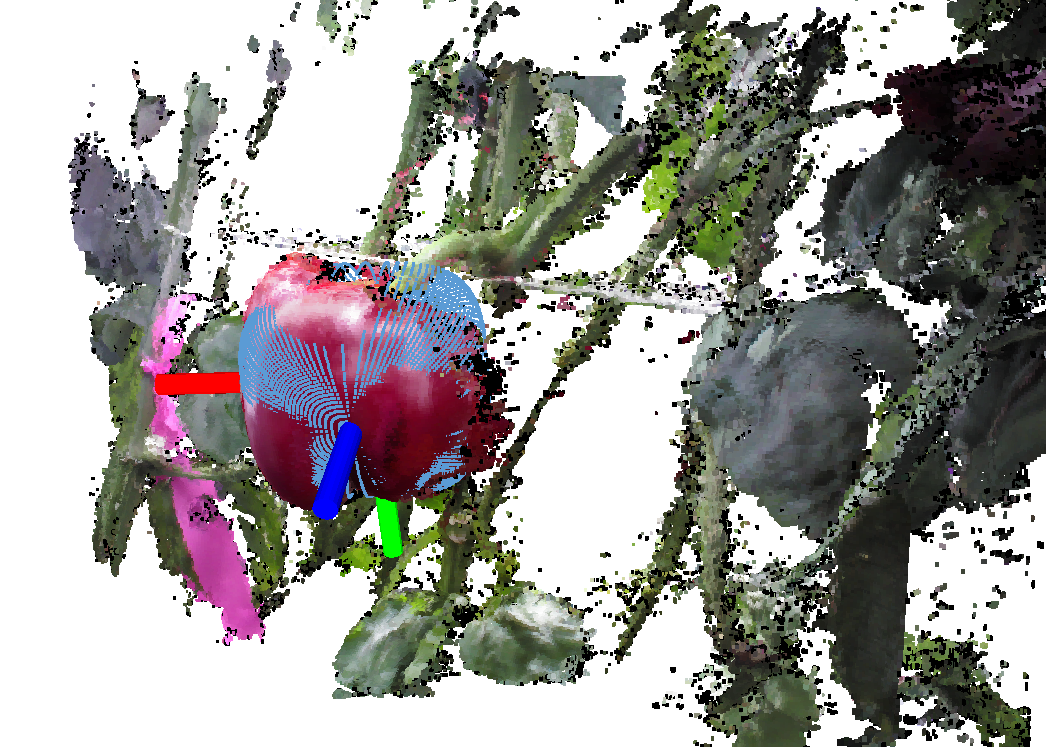}
	\end{subfigure}
	
	\begin{subfigure}[b]{\columnwidth}\centering
		
		\includegraphics[width=0.32\columnwidth]{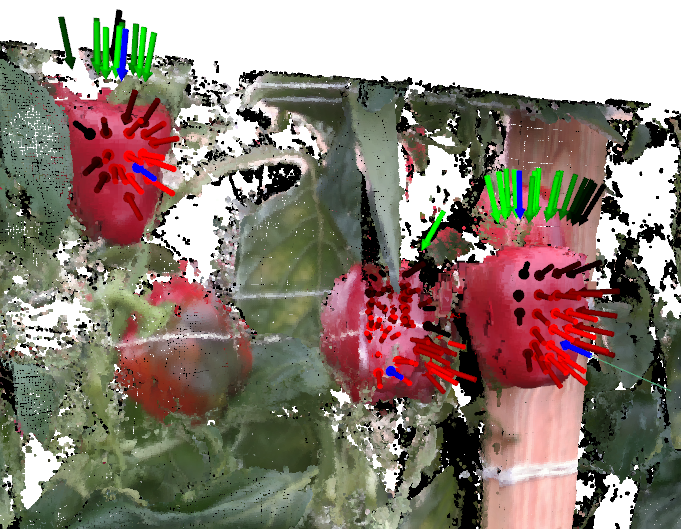}
		\includegraphics[width=0.32\columnwidth]{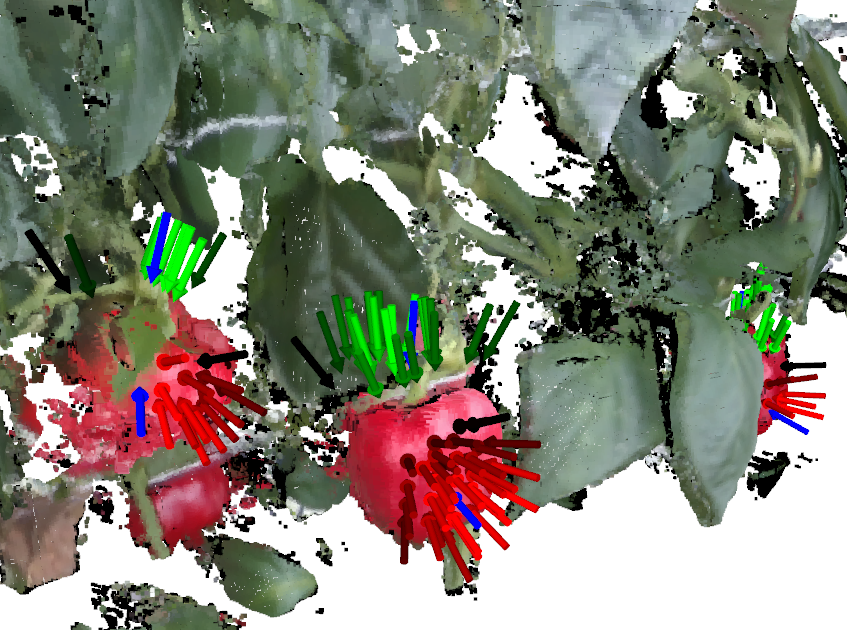}

	\end{subfigure}
	\caption{Reconstructed colour point clouds from scene registration as well as fitted model or selected grasp poses for each detected sweet pepper for 3 and 2 scenes in field trial 1 and 2, respectively.}
	\label{fig:reg_pose_from_field}
	\vspace{-20pt}
\end{figure}

Example sweet peppers harvested by Harvey are shown in Figure \ref{fig:example_sweet_peppers}. An example of a successfully harvested sweet pepper with minor damage (circled in blue) to its top face is shown in Figure \ref{fig:example_sweet_peppers:a}. The model and grasp selection results for five separate trials are also provided in Figure \ref{fig:reg_pose_from_field}. It can be seen qualitatively that, for most cases, the model and grasp selection methods find appropriate locations for grasp points. In some cases, a poor estimate of the model was found and this is believed to be caused by irregularly shaped sweet peppers. An example of a poor estimate can be seen in the top right example within Figure \ref{fig:reg_pose_from_field}.

\definecolor{light_green}{rgb}{0.77,0.93,0.8}
\definecolor{light_yellow}{rgb}{1.0,0.93,0.61}
\definecolor{light_red}{rgb}{1.0,0.78,0.8}
\newcommand{\good}{\cellcolor{light_green}}
\newcommand{\bad}{\cellcolor{light_red}}
\newcommand{\okay}{\cellcolor{light_yellow}}

\begin{table}[!b]
		\vspace{-1pt}
	\centering
	\caption{Sweet Pepper Individual Harvesting Results}
	\begin{tabularx}{\columnwidth}{s|Yssm{1cm}||Ysss}
		& \multicolumn{4}{|c}{Trial 1 - Cultivar: Claire}     & \multicolumn{4}{||c}{Trial 2 - Cultivar: Redjet} \\ \midrule
		\#          & No. Attempts & Attach     & Detach     & Notes           & No. Attempts & Attach     & Detach     & Notes  \\ \midrule
		1           & 1            & \good S    & \good S    & MD         & 6            & \good S    & \good S    &  \\
		2           & 2            & \good S    & \good S    &            & 2            & \good S    & \bad F     &  \\
		3           & 1            & \good S    & \good S    &            & 3            & \good S    & \good S    &  \\
		4           & 4            & \bad F     & \good S    & OC         & 2            & \good S    & \bad F     &  \\
		5           & 4            & \bad F     & \good S    & IS         & 2            & \good S    & \good S    &  \\
		6           & 4            & \bad F     & \bad F     & OB, IS     & 3            & \bad F     & \bad F     &  \\
		7           & 3            & \bad F     & \good S    & MD         & 1            & \good S    & \bad F     & OB      \\
		8           & 3            & \good S    & \good S    & OC         & 6            & \good S    & \bad F     & OB      \\
		9           & 1            & \bad F     & \good S    & OB, IS     & 4            & \good S    & \good S    & XD      \\
		10          & 2            & \bad F     & \good S    &            & 2            & \good S    & \bad F     &  \\
		11          & 1            & \bad F     & \good S    & MD, OB     & 2            & \bad F     & \bad F     &  \\
		12          & 1            & \good S    & \good S    &            & 1            & \bad F     & \bad F     &  \\
		13          & 1            & \good S    & \good S    &            & 1            & \good S    & \good S    &  \\
		14          & 1            & \good S    & \good S    & OC         & 2            & \good S    & \good S    &  \\
		15          & 1            & \good S    & \good S    &            & 2            & \bad F     & \bad F     &  \\
		16          & 1            & \good S    & \good S    &            & 1            & \good S    & \bad F     &  \\
		17          & 2            & \good S    & \good S    & MD         & 2            & \good S    & \good S    & XD      \\
		18          & 2            & \good S    & \good S    &            & 4            & \good S    & \good S    &  \\
		19          & 2            & \bad F     & \bad F     & XD, IS     & 2            & \good S    & \bad F     & XD      \\
		20          & 2            & \bad F     & \bad F     & OC         & 1            & \good S    & \good S    & OB      \\
		21          & 1            & \good S    & \good S    & OB         & 3            & \good S    & \good S    & OB      \\
		22          & 2            & \bad F     & \bad F     & MD, OC, OB & 1            & \good S    & \bad F     &  \\
		23          & 4            & \good S    & \good S    & MD         & 1            & \good S    & \good S    & OB      \\
		24          & 2            & \good S    & \good S    &            & 1            & \good S    & \bad F     & XD      \\
		25          &             \multicolumn{4}{c||}{ N/A }             & 1            & \bad F     & \bad F     & OB      \\
		26          &             \multicolumn{4}{c||}{ N/A }             & 1            & \good S    & \bad F     &  \\ \bottomrule
	\end{tabularx}%
	\caption*{Where F = Failure, S = Success, XD = Major Damage, MD = Minor Damage, IS = Irregular Sweet pepper, OC = Occluded scan and OB = Obstruction with attachment. }
	\label{tab:individual_harvest_result}%

\end{table}%

\section{Discussion \& Conclusion}\label{sec:discussion_and_conclusion}

This paper describes an autonomous crop harvesting system that achieves state-of-the-art results for sweet pepper harvesting in a protected cropping environment. The system is demonstrated using a custom end-effector that uses both a suction gripper and oscillating blade to successfully remove sweet peppers in a protected cropping environment. The high success rate of detachment (92\%) for field trial 1 is a promising result, especially since detachment is regarded as one of the most challenging aspects of the harvesting process. However, further advancements are necessary for a more general and commercially viable system. 

Most failures of field trial 1 occurred within the attachment stage. In particular, obstructions from leaves or string caused 40\% of the attachment failures, while irregularly shaped sweet peppers caused 30\% of the total attachment failures. In contrast, the attachment rates within field trial 2 are much higher. This can be attributed to the difference of grasp selections between field trial 1 and 2 (surface normals instead of the model fitting approach) and the ability to detect attachment failures using a pressure sensor. In some cases within field trial 1 the suction cup collided with the obstructions while approaching the sweet pepper causing the tool to separate prematurely. This problem was avoided within field trial 2 by the addition of a sensor measuring whether premature separation occurred.

The most common detachment failure was found to be the cutting tool missing either side of the peduncle. In this work the cutting point was calculated by assuming the peduncle protrudes vertically from the centre of the sweet pepper. This assumption occasionally breaks since some sweet peppers have peduncles that do not grow vertically (such as the second cultivar Redjet). To improve the detachment reliability, future work will be aimed not only at detecting sweet peppers, but also at detecting the peduncle. We have conducted preliminary work that shows promising results using Point Feature Histograms (PFH) for peduncle detection in \cite{Sa2016}.

The lower detachment success rate for field trial 2 is most likely attributed to the different cultivar (Redjet) involved in the second trial. It was observed that the Redjet cultivar had more challenges associated with it, including shorter, thicker peduncles and more crop growing on the inside of the canopy, increasing the number of obstructions. This result highlights the importance of selecting cultivar that are more suitable for automated harvesting.

The novel contributions of this work have resulted in significant and encouraging improvements in sweet pepper picking success rates compared with the state-of-the-art. The methods presented in this paper provides steps towards the goal of fully autonomous and reliable crop picking systems that will revolutionise the horticulture industry by reducing labour costs, maximising the quality of produce, and ultimately improving the sustainability of farming enterprises.

\bibliography{./bibs/RALRefs,./bibs/mccool_bib}

%

\end{document}